\documentclass[sigconf, nonacm]{acmart}

\AtBeginDocument{%
  }

\usepackage{booktabs}
\usepackage{multirow}
\usepackage{tabularx}
\usepackage{makecell}
\usepackage{array}
\usepackage{float}
\usepackage{amsmath}
\usepackage{graphicx}
\usepackage{xcolor}
\usepackage{listings}
\usepackage[capitalize,noabbrev]{cleveref}
\renewcommand{\arraystretch}{1.12}

\lstdefinestyle{auditpy}{
  language=Python,
  basicstyle=\ttfamily\footnotesize,
  keywordstyle=\color{blue!70!black}\bfseries,
  commentstyle=\color{gray!80!black}\itshape,
  stringstyle=\color{green!50!black},
  numbers=left,
  numberstyle=\tiny\color{gray},
  numbersep=6pt,
  showstringspaces=false,
  breaklines=true,
  frame=single,
  framesep=4pt,
  rulecolor=\color{black!30},
  columns=fullflexible,
  keepspaces=true,
}

\begin{document}

\title{CRAFT: Learn the Schema, Execute the Plan}

\author{Aakash Kolekar}
\authornote{Both authors contributed equally to this research.}
\orcid{0000-0003-3129-7954}
\affiliation{%
  \institution{Amazon Advertising Foundations}
  \city{Palo Alto}
  \state{CA}
  \country{USA}}
\email{aakashvv@amazon.com}

\author{Sahika Genc}
\authornotemark[1]
\affiliation{%
  \institution{Amazon Web Services Agentic AI}
  \city{Seattle}
  \state{WA}
  \country{USA}}
\email{sahika@amazon.com}

\author{Shahriar Shariat}
\affiliation{%
  \institution{Amazon Advertising Foundations}
  \city{Palo Alto}
  \state{CA}
  \country{USA}}
\email{sshariat@amazon.com}

\author{Bunyamin Sisman}
\affiliation{%
  \institution{Amazon Advertising Foundations}
  \city{Seattle}
  \state{WA}
  \country{USA}}
\email{bunyamis@amazon.com}

\author{Tibor Mezi}
\affiliation{%
  \institution{Amazon Advertising Foundations}
  \city{Seattle}
  \state{WA}
  \country{USA}}
\email{meztibor@amazon.com}

\author{Barbara Poblete}
\affiliation{%
  \institution{Amazon Advertising Foundations}
  \city{Seattle}
  \state{WA}
  \country{USA}}
\email{bpoblete@amazon.com}

\author{Shree Vandana Kachroo}
\affiliation{%
  \institution{Amazon Advertising Foundations}
  \city{Seattle}
  \state{WA}
  \country{USA}}
\email{kacshree@amazon.com}

\author{Calvin Chi}
\affiliation{%
  \institution{Amazon Advertising Foundations}
  \city{Seattle}
  \state{WA}
  \country{USA}}
\email{cachi@amazon.com}

\author{Parth Parmar}
\affiliation{%
  \institution{Amazon Advertising Foundations}
  \city{New York}
  \state{NY}
  \country{USA}}
\email{pkparmar@amazon.com}

\author{Ari Singer}
\affiliation{%
  \institution{Amazon Advertising Foundations}
  \city{New York}
  \state{NY}
  \country{USA}}
\email{arising@amazon.com}

\author{Prayaas Jain}
\affiliation{%
  \institution{Amazon Advertising Foundations}
  \city{New York}
  \state{NY}
  \country{USA}}
\email{jainpray@amazon.com}

\author{Cindy Barker}
\affiliation{%
  \institution{Amazon Advertising Foundations}
  \city{San Diego}
  \state{CA}
  \country{USA}}
\email{cindybr@amazon.com}

\author{Benoit Dumoulin}
\affiliation{%
  \institution{Amazon Advertising Foundations}
  \city{Palo Alto}
  \state{CA}
  \country{USA}}
\email{bdumouli@amazon.com}

\renewcommand{\shortauthors}{Kolekar et al.}

\begin{abstract}
Enterprise coding agents translate natural-language analytical requests into executable code over proprietary APIs, schemas, and metric definitions. Yet the prevailing deployment pattern injecting exhaustive schema and tool documentation into each prompt increases inference overhead, complicates schema evolution, and undermines reliability in multi-turn analysis. We investigate whether stable schema knowledge and tool-use behavior can instead be acquired through post-training while preserving the consistency required for production-facing analytics. We present \textbf{CRAFT}, a two-stage post-training recipe for schema-grounded coding agents. First, schema-stripped PLAN supervised fine-tuning learns domain-structured plans and executable behaviors from validated trajectories without exhaustive prompt-time schema injection. Second, execution-shaped reinforcement learning aligns the policy for tool selection, code quality, plan--code consistency, and recovery from failed executions. Training trajectories are curated through a Tri-Gate filter combining execution validation, data-integrity checks, and LLM-judge reasoning audit. We evaluate CRAFT for planned rollout in advertising analytics, covering campaign performance analysis, metric drill-downs, entity-level performance analysis, and multi-turn analytical refinement. The enterprise evaluation environment incorporates beta APIs as the agent-facing tool surface and spans 25 schema-linked core entities and 30 agentic workflows. Relative to a schema-stuffed baseline, CRAFT improves composite Agent Score by +9.6 pp, consistency by +4.1 pp, and multi-turn coherence by +4.2 pp, while reducing input-token burden by approximately 9x and schema-discovery loops by up to 5x. We further report deployment tradeoffs, reward-shaping limitations, and training-infrastructure extensions required for multi-turn tool-use reinforcement learning in enterprise settings.
\end{abstract}



\maketitle

\section{Introduction}
\label{sec:intro}

Code-generating agents are becoming a practical interface for enterprise analytics: users ask natural-language questions, and the agent writes executable code that calls internal APIs, joins structured data, computes metrics, and returns an analysis. In advertising analytics, these agents must operate over proprietary schemas, tool signatures, metric definitions, and domain constraints. The default deployment pattern is \emph{context stuffing}: each prompt includes a large block of Data Definition Language (DDL), API documentation, and tool instructions. This pattern is problematic for applied deployment. First, long schema prompts compete with user turns, tool traces, and intermediate results for working memory~\cite{liu2023lost,he2024neverlost}. Second, every turn repays a large token overhead. Third, physical schema changes such as column renames or table migrations require prompt-pipeline updates and regression testing.

We study CRAFT, a production-oriented coding-agent system evaluated for planned rollout in advertising analytics. The intended users are advertisers, product teams, and subject-matter experts (SMEs) who need natural-language access to analytical workflows such as campaign performance analysis, metric drill-downs, and entity-level comparisons. CRAFT is evaluated in an enterprise evaluation environment that incorporates beta APIs representing the intended production tool surface together with controlled schema-linked advertising data. This setup exercises tool invocation behavior against beta APIs while preserving privacy, repeatability, and isolation from customer-sensitive production data.

The central design decision is to move stable schema and tool-use behavior from transient prompt context into post-training. Stage I performs schema-stripped PLAN supervised fine-tuning (PLAN SFT): demonstrations include business-level plans and executable code, but exhaustive DDL is removed from the input and tool outputs are masked from the loss. Stage II applies execution-shaped reinforcement learning using Group Relative Policy Optimization (GRPO), rewarding correctness, tool-call structure, code quality, plan consistency, and business-aligned judgments.

\paragraph{Applied contributions.}
\begin{enumerate}
\itemsep0em
    \item \textbf{A planned-rollout enterprise analytics system.} We describe the deployment setting, users, constraints, and evaluation methodology for a coding agent intended for advertising analytics workflows.
    \item \textbf{A schema-stripped post-training recipe.} CRAFT combines PLAN SFT and execution-shaped RL to reduce reliance on exhaustive prompt-time schema injection.
    \item \textbf{A production-oriented evaluation.} We evaluate task quality, consistency, multi-turn interaction, inference efficiency, and general-capability guardrails over 25 schema-linked entities and 30 agentic workflows, reporting confidentiality-safe deltas over the same schema-stuffed baseline. A subset of the schema can be obtained by processing publicly available APIs for Amazon Ads in \cite{AmazonAdsAPIv1, AmazonAdsAPIoverview}.
    \item \textbf{Deployment lessons and infrastructure extensions.} We document design tradeoffs, failure modes, reward-shaping limitations, and the VERL/SLIME extensions needed for multi-turn tool-use rollouts.
\end{enumerate}

\begin{figure}[!t]
  \centering
  \includegraphics[width=\columnwidth]{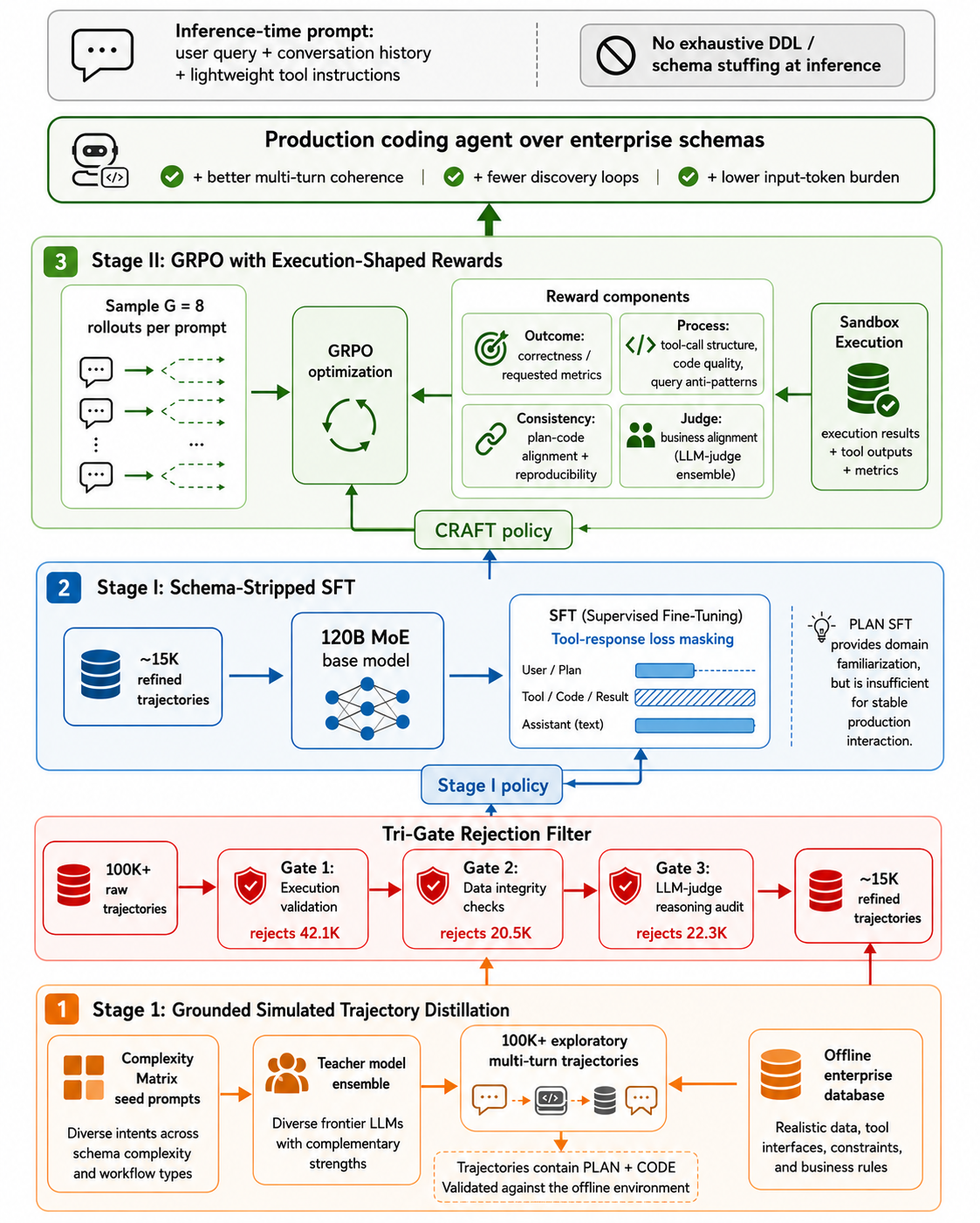}
  \Description{A four-stage pipeline: enterprise-grounded trajectory distillation; Tri-Gate filtering; schema-stripped PLAN supervised fine-tuning; and GRPO with execution-shaped rewards.}
  \caption{CRAFT system overview. In the existing diagram, the offline enterprise database is the controlled schema-linked data component of an evaluation environment whose tool-facing surface incorporates beta APIs. The Tri-Gate filter yields 15.1K accepted trajectories; PLAN SFT removes exhaustive schema context from inference; and execution-shaped GRPO improves interaction behavior.}
  \label{fig:craft_architecture}
\end{figure}

\section{Deployment Context and Requirements}
\label{sec:deployment}

CRAFT is evaluated for planned rollout in advertising analytics workflows used by advertisers, product teams, and subject-matter experts. These workflows require the agent to interpret business intent, identify relevant schema-linked entities and metrics, invoke appropriate tools, execute code, and synthesize an answer that remains consistent across follow-up turns. Representative tasks include campaign performance analysis, metric drill-downs, and entity-level comparisons.

The evaluation is conducted in an enterprise evaluation environment that incorporates beta APIs as part of the intended tool surface and uses controlled schema-linked advertising data for repeatable execution. This design avoids evaluating only against synthetic tool behavior: the agent must issue structured calls through beta APIs while the environment provides privacy-preserving and reproducible analytical outcomes. Customer-sensitive production data is not exposed in training or evaluation artifacts.
Figure~\ref{fig:craft_architecture} labels its controlled data-store component as an ``offline enterprise database;'' beta APIs constitute the tool-facing interface of the broader evaluation environment.

\begin{table*}[!t]
\caption{Planned-rollout setting and pre-deployment evidence. Metrics are reported as deltas or ratios over a fixed schema-stuffed baseline because absolute operational scores and proprietary schema details are confidential. A subset of the schema is publicly available thorough processing for Amazon Ads API definitions in \cite{AmazonAdsAPIv1}.}
\label{tab:deployment_summary}
\centering
\small
\renewcommand{\arraystretch}{1.18}
\begin{tabularx}{\textwidth}{@{}p{0.19\textwidth}p{0.27\textwidth}p{0.20\textwidth}X@{}}
\toprule
\textbf{Dimension} & \textbf{Evidence} & \textbf{Dimension} & \textbf{Evidence} \\
\midrule
Deployment stage & Planned rollout / pre-deployment evaluation & Intended users & Advertisers, product teams, SMEs \\
Application domain & Advertising analytics & Agentic workflow coverage & 25 schema-linked entities (e.g. Advertisers, AdGroups, Ads, etc); 30 workflows \\
Evaluation environment & Controlled data with beta-API tool surface & Training trajectories & 100K+ generated; 15.1K accepted \\
Quality reporting & Deltas over schema-stuffed baseline & Efficiency reporting & Input-token and schema-discovery-loop ratios \\
\bottomrule
\end{tabularx}
\end{table*}

The application has three requirements. \textbf{Reliability}: small metric or schema mistakes can produce incorrect business-facing answers. \textbf{Efficiency}: repeatedly injecting exhaustive schema context creates high token overhead and repeated tool discovery. \textbf{Interaction quality}: users refine requests, add constraints, and ask follow-up questions, so the system must retain context and advance the analysis rather than treating each turn independently.

We use \emph{agentic workflow} to denote an analytical request that cannot be completed by a single static lookup: the agent must select one or more tools, construct executable operations over schema-linked entities, interpret retrieved evidence, and communicate an answer. The planned-rollout scope includes campaign performance analysis, metric drill-downs, entity-level comparisons, and follow-up refinements that change scope or add constraints.

\section{Enterprise-Grounded Trajectory Distillation}
\label{sec:distillation}

A core bottleneck is the lack of human ground truth for long-horizon, multi-tool code generation over proprietary schemas. We treat trajectory generation as a verifiable search problem. Teacher models generate candidate trajectories in the enterprise evaluation environment, which combines beta-API tool interactions with controlled schema-linked advertising data that preserves production-relevant entity relationships, metric constraints, and query paths while masking customer-sensitive values.

\subsection{Coverage and Trajectory Format}

To avoid narrow prompt coverage, we seed generation with a complexity matrix spanning two axes. \emph{Topological breadth} captures the number of entity-graph hops required by a task. \emph{Execution depth} captures filtering, aggregation, metric composition, temporal comparisons, and cross-neighborhood logic. Candidate trajectories use a three-block format: a reasoning scratchpad used during generation, a business-level \texttt{<PLAN>} block explaining entities and metrics without raw DDL, and an executable \texttt{<CODE>} block used for validation.

\begin{table}[!t]
\caption{Tri-Gate filtering. Removed counts are reported because later gates operate only on retained trajectories.}
\label{tab:trigate}
\centering
\footnotesize
\renewcommand{\arraystretch}{1.18}
\begin{tabular}{@{}llr@{}}
\toprule
\textbf{Stage} & \textbf{Failure removed} & \textbf{Count} \\
\midrule
Gate 1 & Execution failure / timeout & 42.1K \\
Gate 2 & Empty / incomplete output & 20.5K \\
Gate 3 & Reasoning mismatch & 22.3K \\
\midrule
Accepted & Validated trajectories & 15.1K \\
\bottomrule
\end{tabular}
\end{table}

\subsection{Tri-Gate Rejection Filter}

Table~\ref{tab:trigate} summarizes the filtering of exploratory generations. Gate 1 rejects non-executable code, including syntax errors, failed API calls, and execution timeouts. Gate 2 rejects outputs that are empty or missing requested target metrics. Gate 3 uses an LLM-judge ensemble~\cite{chen2025multiagent} to detect semantic mismatches between the prompt, plan, and code. The filter turns 100K+ exploratory generations into 15.1K accepted trajectories. We report removed counts rather than per-gate percentages because later gates operate only on trajectories not removed by earlier gates. Applying deterministic checks before semantic judgment ensures that the retained plans are grounded in executable evidence before LLM-judge assessment.

\section{Post-Training System}
\label{sec:method}

\subsection{PLAN SFT with Schema Stripping}

Stage I tests whether the agent can learn reusable schema and API patterns without exhaustive prompt-time DDL. During SFT on a 120B MoE base, the 50K+ token schema context is removed from the input. Tool outputs are masked from the loss because they contain unpredictable identifiers, timestamps, runtime values, and dataframe contents~\cite{schick2023toolformer,patil2023gorilla}. For trajectory tokens $\tau=\{x_1,\ldots,x_T\}$:
\begin{equation}
\mathcal{L}=-\sum_t m_t \log P(x_t\mid x_{<t}),
\end{equation}
where $m_t=1$ for assistant-generated planning and code tokens, and $m_t=0$ for user inputs, system messages, tool outputs, execution stdout, and runtime artifacts. PLAN SFT provides domain familiarity, but our evaluation shows that imitation alone does not fully align the agent's tool-use or ambiguity-resolution policy.

\subsection{Execution-Shaped GRPO}

Stage II uses GRPO~\cite{shao2024deepseekmath} to optimize behavior under execution feedback. We sample $G=8$ trajectories per prompt and optimize:
\begin{equation}
J(\theta)=\mathbb{E}_{q,o}\left[R_{\text{total}}(q,o)-\beta\,\mathrm{KL}(\pi_\theta\|\pi_{\mathrm{ref}})\right],
\end{equation}
with:
\begin{equation}
R_{\text{total}}=R_{\text{outcome}}+\alpha R_{\text{process}}+\lambda R_{\text{consistency}}+\gamma R_{\text{judge}}.
\end{equation}

\begin{table}[!t]
\caption{Reward components. Execution remains the primary grounding signal.}
\label{tab:reward_components}
\centering
\footnotesize
\renewcommand{\arraystretch}{1.18}
\begin{tabularx}{\columnwidth}{@{}p{0.25\columnwidth}p{0.34\columnwidth}X@{}}
\toprule
\textbf{Reward} & \textbf{Signal} & \textbf{Purpose} \\
\midrule
$R_{\text{outcome}}$ & Execution / unit tests & Correctness \\
$R_{\text{process}}$ & Tool-call structure; code quality & Behavior shaping \\
$R_{\text{consistency}}$ & Plan-code / run agreement & Stability \\
$R_{\text{judge}}$ & LLM-judge assessment & Business alignment \\
\bottomrule
\end{tabularx}
\end{table}

Table~\ref{tab:reward_components} summarizes the reward. $R_{\text{process}}$ rewards well-structured tool use, valid tool-call sequencing, clear code organization, and avoidance of obvious query anti-patterns such as redundant joins or broad scans. It does not yet fully optimize low-level memory footprint or infrastructure-aware runtime cost; this limitation is discussed in \cref{sec:limitations}. Appendix~\ref{app:judge_safeguards} describes how judge-based signals are bounded by execution evidence, and Appendix~\ref{app:infra} describes the VERL/SLIME extensions used to propagate step-level tool rewards and support asynchronous multi-turn rollouts. This design intentionally prioritizes correct and stable analytical behavior before attempting to optimize heterogeneous infrastructure-level costs.

\section{Evaluation}
\label{sec:evaluation}

We conduct a pre-deployment evaluation over advertising analytics tasks spanning 25 schema-linked core entities and 30 agentic workflows. The enterprise evaluation environment incorporates beta APIs as the agent-facing tool surface and controlled schema-linked data for reproducible execution. The evaluation measures task-level quality, consistency, multi-turn interaction, efficiency, and general-capability guardrails. Unless noted, reported values are means over three seeded runs. Because schemas, APIs, and evaluation rubrics are confidential, production metrics are reported as deltas over the same schema-stuffed baseline; efficiency metrics are reported as ratios where the baseline is 1.0x.

\paragraph{Evaluation layers.}
\textbf{Layer 1: Capability isolation.} We use public benchmarks and internal skill checks without agentic scaffolding to assess whether domain adaptation preserves general reasoning and instruction following. \textbf{Layer 2: Beta-API tool-use evaluation.} The full agent stack executes agentic workflows through beta APIs within the enterprise evaluation environment; we score tool selection, argument validity, data retrieval, analysis, and final-answer correctness. \textbf{Layer 3: Consistency.} Repeated sampling measures whether tool-use decisions and final answers remain semantically equivalent across runs. \textbf{Layer 4: Multi-turn interaction.} Realistic conversations test clarification, follow-up constraints, pivots, and recovery from execution failure.

Appendix~\ref{app:code_baselines} supports the schema-stuffed baseline comparison; Appendix~\ref{app:failure_taxonomy} describes observed failure categories; Appendix~\ref{app:judge_safeguards} documents the safeguards around judge-based signals; Appendix~\ref{app:infra} summarizes training-gym extensions; and Appendix~\ref{app:gsm8k_audit} supports the public-benchmark guardrail analysis.

\subsection{Evaluation Tasks and Metric Construction}
\label{sec:metric_construction}

The evaluation tasks reflect analytical interactions expected in the planned rollout setting rather than isolated code-completion problems. They include campaign performance analysis, metric drill-downs, cross-entity aggregation, ambiguous-scope requests that require clarification, multi-turn refinements that add constraints or change entity scope, and recovery from incomplete or failed tool executions. These task families test whether an agent identifies an appropriate analytical path and preserves user intent, in addition to whether its code executes. This emphasis is related to conversational database interaction in CoSQL~\cite{yao2019cosql} and enterprise-scale analytical workflow evaluation in Spider~2.0~\cite{lei2025spider2}, while our setting evaluates proprietary beta-API interactions under confidentiality-preserving reporting.

We summarize task-level quality with a composite Agent Score:
\begin{equation}
\mathrm{AgentScore}=w_{\mathrm{exec}}S_{\mathrm{exec}}+w_{\mathrm{tool}}S_{\mathrm{tool}}+w_{\mathrm{plan}}S_{\mathrm{plan}}+w_{\mathrm{ans}}S_{\mathrm{ans}},
\end{equation} 
where $S_{\mathrm{exec}}$ measures successful execution and requested-metric retrieval, $S_{\mathrm{tool}}$ measures appropriate beta-API tool selection and valid argument structure, $S_{\mathrm{plan}}$ measures alignment between the stated analytical plan and generated code, and $S_{\mathrm{ans}}$ measures whether the final response faithfully communicates the computed result. The same fixed scoring rubric is used across compared systems; component deltas are reported in Table~\ref{tab:agent_score_component} so the aggregate improvement can be interpreted rather than treated as a single opaque score.

Interaction metrics target failure modes that matter in a user-facing analytical assistant. \emph{Consistency} measures whether repeated runs resolve scope and choose tool paths compatibly. In multi-turn interactions, \emph{coherence} measures whether later responses remain logically consistent with the analytical thread; \emph{retention} measures preservation of previously supplied constraints; and \emph{progression} measures whether the agent advances the task instead of restarting or repeating earlier work. \emph{Schema-discovery loops} count repeated exploration rounds before reaching a successful executable path, capturing avoidable friction even when a final answer is eventually produced.

All systems use the same held-out tasks, beta APIs, controlled execution data, sandbox, judge rubrics, and decoding settings. Schema-stuffed baselines receive exhaustive DDL/API context; CRAFT receives only the user query, conversation history, and lightweight tool instructions. General-capability benchmarks are used only as guardrails against material regression after adaptation.

\subsection{Training-Stage Ablation}

\begin{table}[!t]
\caption{Training-stage ablation. Deltas are relative to the schema-stuffed baseline; positive values are better.}
\label{tab:progression}
\centering
\footnotesize
\renewcommand{\arraystretch}{1.18}
\begin{tabular}{@{}lccc@{}}
\toprule
\textbf{Stage} & \textbf{Agent} & \textbf{Consist.} & \textbf{Coher.} \\
 & \textbf{Score $\Delta$} & \textbf{$\Delta$} & \textbf{$\Delta$} \\
\midrule
I: PLAN SFT & +2.6 $\pm$ 0.7 & -1.4 $\pm$ 0.5 & +0.8 $\pm$ 0.9 \\
\textbf{II: PLAN SFT + RL} & \textbf{+9.6 $\pm$ 0.4} & \textbf{+4.1 $\pm$ 0.3} & \textbf{+4.2 $\pm$ 0.6} \\
\bottomrule
\multicolumn{4}{@{}l}{\scriptsize All values in percentage points (pp).}
\end{tabular}
\end{table}

Table~\ref{tab:progression} shows that PLAN SFT provides modest domain familiarity but slightly reduces consistency. This pattern is informative: removing exhaustive schema context asks the model to rely on internalized domain structure, but imitation learning alone does not determine how reliably the agent should resolve ambiguous scopes, choose between alternative tool paths, or recover when an initial analysis is incomplete. PLAN SFT should therefore be interpreted as a necessary initialization rather than the final interaction policy.

\begin{table}[!t]
\caption{Relative inference efficiency. The schema-stuffed baseline is 1.00x; lower is better.}
\label{tab:efficiency}
\centering
\footnotesize
\renewcommand{\arraystretch}{1.18}
\begin{tabular}{@{}lcc@{}}
\toprule
\textbf{System} & \textbf{Input} & \textbf{Schema-Discovery} \\
 & \textbf{Tokens} & \textbf{Loops} \\
\midrule
Baseline + DDL & 1.00x & 1.00x \\
Stage I: PLAN SFT & 0.11x & 0.62x \\
\textbf{Stage II: PLAN SFT + RL} & \textbf{0.11x} & \textbf{0.20x} \\
\bottomrule
\end{tabular}
\end{table}

\begin{table*}[!t]
\caption{Multi-turn interaction-quality deltas relative to the same schema-stuffed baseline. Positive values indicate improved user-facing interaction behavior.}
\label{tab:multiturn_interaction}
\centering
\small
\renewcommand{\arraystretch}{1.22}
\begin{tabular}{lccccc}
\toprule
\textbf{Training condition} & \textbf{Overall $\Delta$} & \textbf{Completion $\Delta$} & \textbf{Coherence $\Delta$} & \textbf{Retention $\Delta$} & \textbf{Progression $\Delta$} \\
\midrule
Stage I: PLAN SFT & +2.4 pp & +0.0 pp & +0.8 pp & +3.6 pp & +4.5 pp \\
\textbf{Stage II: PLAN SFT + RL} & \textbf{+6.1 pp} & \textbf{+0.5 pp} & \textbf{+4.2 pp} & \textbf{+7.8 pp} & \textbf{+9.4 pp} \\
\bottomrule
\end{tabular}
\end{table*}

The efficiency results in Table~\ref{tab:efficiency} clarify the division of labor between stages. PLAN SFT accounts for essentially all of the input-token reduction: after schema stripping, both Stage I and Stage II operate at 0.11x of the schema-stuffed baseline's input-token burden. In contrast, the dominant reduction in schema-discovery loops arises after execution-shaped RL, falling from 0.62x after PLAN SFT to 0.20x after Stage II. Together with the consistency and coherence gains, this indicates that RL is not merely improving answer quality; it is aligning the agent's interactive decision policy for tool selection, ambiguity resolution, and recovery across turns.

Table~\ref{tab:efficiency} reports that removing exhaustive schema context reduces input-token burden by approximately 9x. RL further reduces discovery loops to 0.20x of the schema-stuffed baseline. Table~\ref{tab:multiturn_interaction} shows that the largest multi-turn gains appear in retention and progression. This is important for applied usage: advertisers and product teams refine goals, add constraints, and expect the agent to maintain an analytical thread. Agent Score is a weighted combination of execution success, tool-use correctness, plan-code alignment, and final-answer accuracy. Table~\ref{tab:agent_score_component} shows that improvements are spread across components rather than concentrated in a single axis.

\begin{table*}[!t]
\caption{Agent Score component deltas relative to the same schema-stuffed baseline. Stage II improves all scored components rather than optimizing only execution success.}
\label{tab:agent_score_component}
\centering
\small
\renewcommand{\arraystretch}{1.22}
\begin{tabular}{lcccc}
\toprule
\textbf{Training condition} & \textbf{Execution $\Delta$} & \textbf{Tool-use $\Delta$} & \textbf{Plan-code $\Delta$} & \textbf{Final-answer $\Delta$} \\
\midrule
Stage I: PLAN SFT & +2.3 pp & +0.8 pp & +2.5 pp & +1.6 pp \\
\textbf{Stage II: PLAN SFT + RL} & \textbf{+7.1 pp} & \textbf{+2.8 pp} & \textbf{+8.4 pp} & \textbf{+6.2 pp} \\
\bottomrule
\end{tabular}
\end{table*}

\begin{table}[!t]
\caption{General-capability guardrails reported as deltas over the same pre-adaptation base model.}
\label{tab:guardrails}
\centering
\footnotesize
\renewcommand{\arraystretch}{1.18}
\begin{tabular}{@{}lccc@{}}
\toprule
\textbf{Model} & \textbf{IFEval} & \textbf{GSM8K} & \textbf{GPQA} \\
\midrule
Larger reference & +0.8 pp & +1.2 pp & -0.7 pp \\
\textbf{CRAFT} & \textbf{+0.6 pp} & \textbf{+0.9 pp} & \textbf{-0.3 pp} \\
\bottomrule
\end{tabular}
\end{table}

Public benchmarks are not the primary target; Table~\ref{tab:guardrails} shows near-zero deltas, indicating no material regression. Appendix~\ref{app:code_baselines} gives the confidentiality-safe comparison against schema-stuffed code-agent baselines, and Appendix~\ref{app:gsm8k_audit} describes the GSM8K contamination audit.

\subsection{Planned Rollout, Release Gates, and Reproducibility}
\label{sec:rollout}
\label{sec:monitoring}

CRAFT is evaluated as a pre-deployment system for planned rollout rather than as an already launched service. The purpose of this evaluation is therefore to support release decisions, not only to compare model variants. Before broader exposure to intended users, we consider four classes of evidence. \emph{Analytical correctness} is measured through successful execution, requested-metric retrieval, and final-answer fidelity. \emph{Interaction reliability} is measured through repeated-run consistency and multi-turn retention and progression. \emph{Tool-use validity} is tested through beta-API invocation behavior, execution failures, and schema-discovery loops. Finally, \emph{capability preservation and generalization} is assessed using guardrail benchmarks after domain adaptation.

This structure is motivated by the failure taxonomy in Appendix~\ref{app:failure_taxonomy}: a tool call can execute successfully while still resolving an ambiguous metric scope incorrectly or losing a previously stated constraint. Consequently, execution and answer failures are release-blocking, while elevated discovery loops, reduced retention, or disagreement-heavy trajectories trigger further rubric review, trajectory curation, or reward-shaping refinement prior to rollout. This separates apparent executability from stable business-facing analytical behavior.

The proprietary schemas, beta APIs, and production-derived evaluation rubrics cannot be released. If accepted, we intend to provide sufficient reproduction instructions for the disclosed VERL/SLIME training-gym extensions and for the structure of the multi-turn rollout, reward propagation, and evaluation pipeline. This enables methodological reproduction on compatible structured-data environments without exposing privileged interfaces or customer-sensitive data.

\section{Design Tradeoffs and Lessons Learned}
\label{sec:tradeoffs}

\textbf{Why not retrieval-only schema grounding?}
Retrieval-augmented generation~\cite{lewis2020rag,sun2024productrag} can efficiently surface a small schema or metadata neighborhood for sparse lookup, but many advertising analytics tasks require multi-hop joins, metric compatibility checks, and follow-up reasoning. CRAFT still uses lightweight tool instructions at runtime, but avoids repeatedly injecting exhaustive schema context.

\textbf{Why PLAN SFT before RL?}
RL from a general-purpose coding model wastes rollout budget on invalid syntax, invalid joins, and hallucinated tools. PLAN SFT narrows the action space by teaching common entity relationships and code idioms. However, Table~\ref{tab:progression} shows that SFT alone is not sufficient; interaction policy alignment requires execution feedback.

\textbf{Why relative reporting?}
Production scores, schemas, and APIs are confidential. Relative deltas and ratios let us quantify impact while avoiding disclosure of sensitive operational details.

\textbf{Why a controlled environment with beta APIs?}
Directly training and debugging against customer production data would create privacy and reliability risks. The enterprise evaluation environment exercises the intended beta-API tool surface while using controlled schema-linked data that preserves analytical constraints and query paths. This provides realistic pre-rollout evidence without exposing privileged values.

\section{Related Work}
\label{sec:related}

\textbf{Structured-data agents and enterprise analytical workflows.}
Natural-language interfaces to structured data have progressed from table and semantic-parsing systems such as TAPAS~\cite{herzig2020tapas}, Spider~\cite{yu2018spider}, RAT-SQL~\cite{wang2020ratsql}, and CoSQL~\cite{yao2019cosql} to LLM-based database-grounded settings such as DIN-SQL~\cite{pourreza2024dinsql}, BIRD~\cite{li2023bird}, and semantic test-suite evaluation~\cite{zhong2020testsuite}. More recently, Spider~2.0~\cite{lei2025spider2} highlights that realistic enterprise workflows require models to navigate large metadata surfaces, documentation, heterogeneous database systems, and multi-step analytical programs. CRAFT targets a related applied challenge over proprietary advertising-analytics interfaces: reducing repeated prompt-time presentation of stable schemas while preserving reliable tool-mediated analysis.

\textbf{Tool-augmented language agents and execution feedback.}
Toolformer~\cite{schick2023toolformer} and Gorilla~\cite{patil2023gorilla} study API invocation by language models, while ReAct~\cite{yao2022react}, Reflexion~\cite{shinn2023reflexion}, and CodeAct~\cite{wang2024codeact} interleave reasoning, actions, and executable code. ToolLLM~\cite{qin2024toolllm}, API-Bank~\cite{li2023api}, and AgentBench~\cite{liu2023agentbench} broaden evaluation toward interactive tool-use behavior. CRAFT differs in focusing on a planned-rollout analytical system where schema-grounded interaction, beta-API invocation, and multi-turn consistency are jointly evaluated.

\textbf{Post-training and execution-grounded optimization.}
Tool-use SFT commonly masks unpredictable tool outputs from supervised loss~\cite{schick2023toolformer,patil2023gorilla}. Recent reasoning-oriented post-training methods use reinforcement learning and execution-verifiable feedback to improve policy behavior beyond imitation~\cite{shao2024deepseekmath,guo2025deepseekr1,wang2024offlinereinforcementlearningllm,yoshihara2025practicaltwostagerecipemathematical}. CRAFT instantiates this separation operationally: PLAN SFT installs schema- and workflow-level familiarity, while execution-shaped RL aligns tool selection, plan--code consistency, recovery behavior, and multi-turn progression.

\textbf{Retrieval, answerability, and judge-based evaluation.}
Retrieval-augmented generation~\cite{lewis2020rag} provides an alternative to placing exhaustive knowledge in every prompt, and product-aware RAG~\cite{sun2024productrag} demonstrates grounding generation in structured commercial metadata. TAONA~\cite{yan2025taona} emphasizes answerability and abstention in graph-grounded question answering, a related concern when analytics agents face ambiguous entity or metric scopes. Because CRAFT uses judge-based signals for reasoning audit and business alignment, it is also informed by work characterizing LLM-as-judge reliability and rubric-based evaluation~\cite{zheng2023judging,kim2023prometheus,chen2025multiagent}. We retain execution and data-integrity checks as primary grounding signals and use judge-based signals only for dimensions difficult to express through deterministic tests.

\section{Operational Safeguards for Planned Rollout}
\label{sec:operational_safeguards}

The planned rollout introduces safeguards at three boundaries. \textbf{Data boundary:} customer-sensitive production values are excluded from training artifacts; controlled schema-linked data enables repeated execution and debugging without exposing privileged records. \textbf{Tool boundary:} beta APIs are used as the agent-facing surface so that tool selection, argument formation, and error handling can be evaluated under interfaces representative of intended deployment. \textbf{Interaction boundary:} ambiguous requests, conflicting metric scopes, and high-disagreement outputs are candidates for clarification or subject-matter-expert review rather than unconditional answer generation. These safeguards complement the quantitative release criteria in Section~\ref{sec:monitoring}: task failures and answer errors are blocking, while elevated discovery loops or degraded retention identify system refinements required before broader exposure.

\section{Limitations}
\label{sec:limitations}

\textbf{Schema-novel entities.} CRAFT targets stable enterprise schemas. Queries that require genuinely new abstract entities or workflows not represented in training may require a fine-tuning refresh or retrieval augmentation.

\textbf{Long-horizon analytical chains.} Interaction quality may degrade as conversations accumulate constraints, results, and pivots. Extending the system with durable interaction memory is a promising direction.

\textbf{Reward-shaping coverage.} The current $R_{\text{process}}$ emphasizes tool-call structure, code quality, and avoidance of obvious query anti-patterns. It does not yet fully optimize low-level memory footprint, physical query planning, or infrastructure-aware runtime cost; future work should incorporate richer execution telemetry and normalized cost signals.

\textbf{Judge-based signals.} LLM-judge feedback may be vulnerable to reward hacking or stylistic over-optimization. Execution outcomes remain the primary grounding signal, while judge-based terms are used only for dimensions such as business alignment and plan consistency. Appendix~\ref{app:judge_safeguards} describes the safeguards intended for rollout evaluation.

These limitations bound CRAFT's claim: it is a post-training recipe for coding agents operating over stable proprietary schemas in a planned-rollout setting, not a replacement for retrieval or in-context tool grounding in arbitrary open-world domains.

\section{Conclusion}

We presented CRAFT, a production-oriented post-training recipe evaluated for planned rollout in advertising analytics. By combining enterprise-grounded trajectory distillation, schema-stripped PLAN SFT, and execution-shaped RL, CRAFT reduces reliance on prompt-time schema injection while improving coding-agent quality, consistency, and multi-turn interaction behavior. Evaluation through beta APIs within a controlled enterprise environment provides realistic pre-deployment evidence under confidentiality constraints.

\appendix

\section{Baseline Comparison}
\label{app:code_baselines}

Table~\ref{tab:code_baselines_appendix} compares CRAFT against the strongest schema-stuffed baseline. We retain the confidentiality-safe reporting convention used throughout the paper.

\begin{table}[H]
\caption{Comparison against the strongest schema-stuffed baseline.}
\label{tab:code_baselines_appendix}
\centering
\footnotesize
\renewcommand{\arraystretch}{1.18}
\begin{tabular}{@{}lccc@{}}
\toprule
\textbf{Method} & \textbf{Exec. Acc.} & \textbf{Pass@1} & \textbf{Input Tok.} \\
\midrule
Schema-stuffed baseline & 0.0 pp & 0.0 pp & 1.00x \\
\textbf{CRAFT} & \textbf{+7.1 pp} & \textbf{+6.2 pp} & \textbf{0.11x} \\
\bottomrule
\end{tabular}
\end{table}

\section{Failure-Mode Taxonomy}
\label{app:failure_taxonomy}

Table~\ref{tab:error_taxonomy} groups diagnostic priorities observed during repeated-run and multi-turn analysis. The ordering motivates the consistency and interaction-focused metrics in the main evaluation without disclosing absolute evaluation-set counts.

\begin{table}[!t]
\caption{Failure taxonomy ordered by diagnostic priority; no absolute evaluation-set counts are disclosed.}
\label{tab:error_taxonomy}
\centering
\footnotesize
\renewcommand{\arraystretch}{1.12}
\begin{tabularx}{\columnwidth}{@{}p{0.34\columnwidth}X@{}}
\toprule
\textbf{Category} & \textbf{Primary failure mode} \\
\midrule
Ambiguity resolution & Metric, scope, or data-source interpretation \\
Behavioral instability & Tool-use or strategy varies across runs \\
Hierarchy / aggregation & Incorrect scope or aggregation level \\
Post-retrieval reasoning & Calculation or labeling mistake \\
Data normalization & Case or string-matching mismatch \\
Tool execution & Invalid connection or invocation \\
Environment stability & Controlled state differs across runs \\
Missing-data handling & Null values interpreted inconsistently \\
\bottomrule
\end{tabularx}
\end{table}

\section{Judge-Based Signal Safeguards}
\label{app:judge_safeguards}

Prior work documents both the usefulness and the potential biases of LLM-as-judge evaluation~\cite{zheng2023judging,kim2023prometheus}. Accordingly, CRAFT uses judge-based signals only for dimensions that are difficult to express through deterministic execution tests, such as business alignment and plan--code consistency. Execution outcomes and data-integrity checks remain the primary grounding signals. To limit stylistic reward gaming, the rollout protocol uses outcome anchoring: trajectories that fail execution cannot receive a favorable aggregate reward solely from judge scores. Prior to broader rollout, high-reward and disagreement cases are candidates for subject-matter-expert review and rubric refinement.

\section{VERL/SLIME Training-Gym Extensions}
\label{app:infra}

The training gym extends VERL~\cite{verl2025} and SLIME~\cite{slime2025} at the rollout and reward layers to support multi-turn tool-use trajectories through the beta-API surface. VERL's rollout output is augmented with an \texttt{extra\_fields} dictionary that carries per-turn tool rewards and judge sub-scores into the custom reward function rather than collapsing supervision into a single terminal scalar. SLIME is extended with a multi-turn rollout generator that alternates model actions, structured tool dispatch, environment observations, and loss-mask construction. We also decouple rollout generation, reward scoring, and training into asynchronous phases so updates do not block on the slowest long trajectory. The GRPO loss, advantage estimation, and KL penalty remain unchanged; the extensions implement the enterprise interaction gym rather than a new RL objective. Structurally, the changes expose intermediate tool outcomes to reward computation, support stateful tool-response turns with masked observations, and overlap generation with reward scoring for trajectories of varying length.

\section{GSM8K Contamination Audit}
\label{app:gsm8k_audit}

We run a 13-gram overlap audit between GSM8K test examples and the retained CRAFT training corpus. Flagged overlaps are manually inspected to distinguish copied problem content from common arithmetic phrasing. Under this protocol, the near-zero GSM8K delta in Table~\ref{tab:guardrails} is interpreted as guardrail evidence rather than evidence of benchmark optimization.

\bibliographystyle{ACM-Reference-Format}
\bibliography{example_paper}

\section*{GenAI Usage Disclosure}
The authors used generative AI tools to assist with language editing and LaTeX formatting. All technical claims, evaluation results, citations, and conclusions were contributed, reviewed and verified by the authors.

\end{document}